\documentclass[10pt,twocolumn,letterpaper]{article} 

\usepackage{avss}
\usepackage{times}
\usepackage{epsfig}
\usepackage{graphicx}
\usepackage{amsmath}
\usepackage{amssymb}
\usepackage{paralist}

\avssfinalcopy 

\begin{document}

\title{Weakly supervised training of deep convolutional neural networks \\ for overhead pedestrian localization in depth fields}

\author{Alessandro Corbetta\\
Department of Applied Physics\\
Eindhoven University of Technology\\
{\tt\small a.corbetta@tue.nl}
\and
Vlado Menkovski\\
Department of Mathematics and  Computer Science\\
Eindhoven University of Technology\\
{\tt\small v.menkovski@tue.nl}
\and
Federico Toschi\\
Department of Applied Physics and\\
Department of Mathematics and Computer Science\\
Eindhoven University of Technology\\
{\tt\small f.toschi@tue.nl}
}

\maketitle

\begin{abstract}
Overhead depth map measurements capture sufficient amount of information to enable human experts to track pedestrians  accurately. However, fully automating this process using image analysis algorithms can be challenging. Even though hand-crafted image analysis algorithms are successful in many common cases, they fail frequently when there are complex interactions of multiple objects in the image. Many of the assumptions underpinning the hand-crafted solutions do not hold in these cases and the multitude of exceptions are hard to model precisely. Deep Learning (DL) algorithms, on the other hand, do not require hand crafted solutions and are the current state-of-the-art in object localization in images. However, they require exceeding amount of annotations to produce successful models. In the case of object localization these annotations are difficult and time consuming to produce. In this work we present an approach for developing pedestrian localization models using DL algorithms with efficient weak supervision from an expert. We circumvent the need for annotation of large corpus of data by annotating only small amount of patches and relying on synthetic data augmentation as a vehicle for injecting expert knowledge in the model training. This approach of weak supervision through expert selection of representative patches, suitable transformations and synthetic data augmentations enables us to successfully develop DL models for pedestrian localization efficiently.
\end{abstract}

\section{Introduction}\label{sect:intro}
Depth field data encodes the distance between each recorded point and the camera plane. This allows for highly-accurate crowd dynamics analyses in real-world scenarios, i.e. outside of laboratory environments~\cite{DBLP:journals/ijon/BoltesS13}. With this technology, for the first time, the ability to analyze between several thousands to few millions actual pedestrian trajectories has been achieved~\cite{Brscic201477,corbetta2016fluctuations,corbetta2016continuous}. This enabled new statistical insights, unbiased by artificial laboratory conditions (e.g. need for participants to wear tracking hats or vests, and dynamics regulated by the experimenter instructions)~\cite{brscic2013person,corbetta2014TRP,seer2014kinects,PhysRevE.89.012811}. Furthermore, depth measurements do naturally protect the privacy of the pedestrians, since individuals remain unrecognizable. This is a requirement for real-life measurements, and a challenge for methods that use imaging rather than depth field data~\cite{johansson2007specification}. 

Pedestrian positioning from depth field data, acquired from sensors such as Microsoft Kinects~\cite{Kinect}, requires addressing two key tasks, namely: background subtraction and head localization. For a number of common cases these two tasks have a straight-forward solution. Since the camera takes a birds eye view, the background can be simply subtracted by removing all points beyond a depth threshold. The head localization can be approached similarly, in fact the points closest to the camera are part of the pedestrian heads. So far, these tasks have been tackled via hand-crafted approaches that rely on expert-tweaked depth-cloud clustering algorithms~\cite{seer2014kinects} (CL). These approaches segment the different objects mainly based on the assumption that a cluster of neighboring points forms a pedestrian. This assumption typically holds when the dynamics on the scene involve low pedestrian densities (about $1.5\,$ped./m$^2$ max~\cite{corbetta2016continuous}), assorted homogeneously (i.e. composed of adults of similar size, with no elements such as strollers, carts, and bikes) and developing in  simple geometric settings (corridors in~\cite{corbetta2016continuous,seer2014kinects,PhysRevE.89.012811}). 

However, such designed approaches do not generalize well and are sensitive to a number of special case scenarios. This can be as simple as a raised hand that is interpreted as a head, but it becomes a larger issue when pedestrian density increases and CL algorithms experience difficulties in disentangling the individuals. Furthermore, these problems only increase when there are other (i.e. non-pedestrian) objects in the scene, such as moving doors, trolleys and obstacles. 
\begin{figure}[t]
  \begin{center}
    \includegraphics[width=0.95\linewidth]{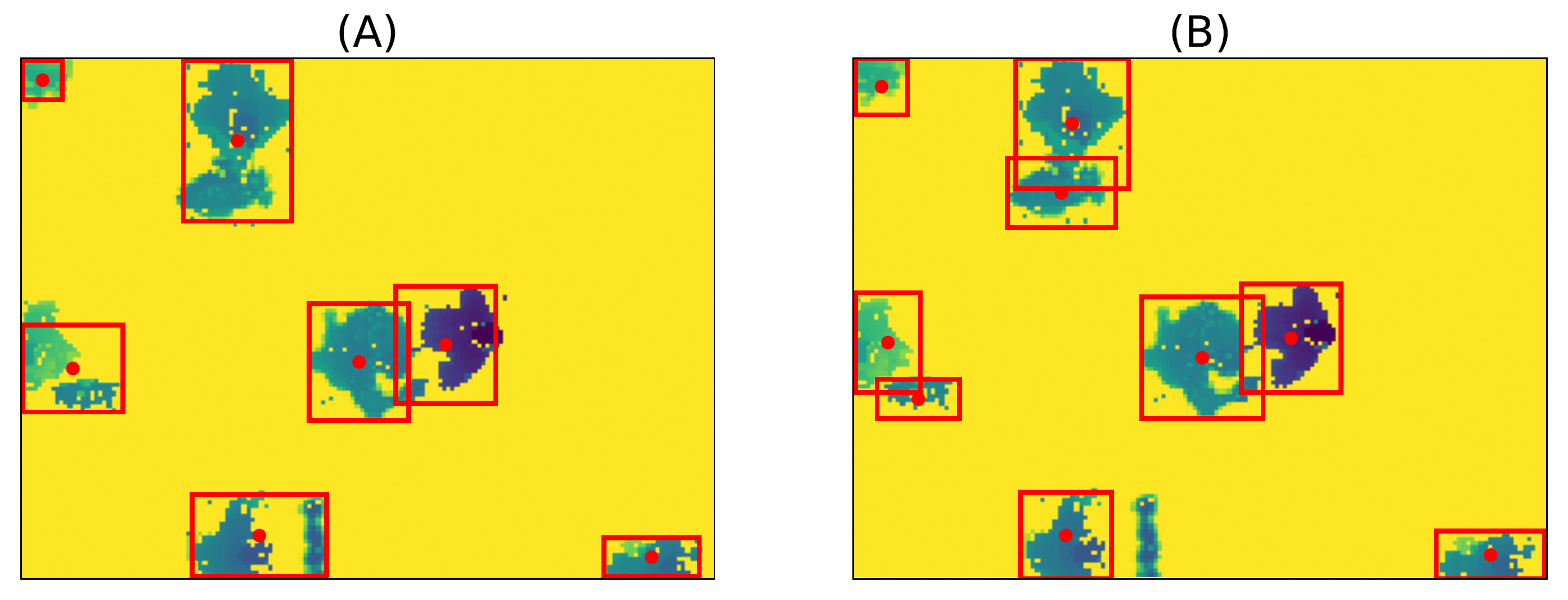}
  \end{center}
  \caption{(A) Example scenario of under-performing localization employing a clustering (CL) approach. The comparison with the ground truth in (B) shows some typical mistakes, e.g. the inability to disentangle close subjects, especially when they are small in size (as infants, cf. top and left of the scene). Furthermore,  typical CL approaches cannot distinguish shapes, thus the localization at the bottom includes a static obstacle.}
  \label{fig:pedClust}
\end{figure}
Figure~\ref{fig:pedClust} contains a sample depth-field map picturing 9 pedestrians, 3 of which are infant (notice the smaller size), and 1 static object. The CL based localization in Fig.~\ref{fig:pedClust}(A) shows some  typical mistakes (cf. ground truth in Fig.~\ref{fig:pedClust}(B)): twice a couple formed by an adult and a child (top and left side of the scene) is detected as a unique individual. Based on the shape of the objects, an  expert instead recognizes that the likely scenario involves two individuals standing close to each other.  The CL algorithms also fails to disentangle a pedestrian at the bottom from a static obstacle in their neighborhood (even though they are not connected). This follows from the small size of the object and the lack of ability of the CL approach to classify shapes. Detailing every exception and crafting appropriate rules to deal with such complexity becomes difficult and requires significant effort that does not transfer well across different measurement scenarios and implementations. 

On the other hand, Machine Learning (ML) methods for image analysis rely on training data rather than design of rules and exceptions. Given sufficient data and good features, these models tend to perform well and are robust to special cases. The success of these approaches has been particularly demonstrated with recent developments in  Deep Learning (DL). DL models obtain excellent performance and are currently state-of-the-art in image localization~\cite{ILSVRC15}. The major advantage that DL methods have over ML image analysis is that they incorporate automatic feature extraction (or representation learning) as part of the model training. However,  one of the main disadvantages of these approaches is the difficulty in incorporating expert knowledge about the problem and hence the requirement of significant amount of annotations~\cite{yu2015lsun}. 
 
Therefore, to develop DL models for pedestrian localization, an expert needs to produce a large amount of hand-annotated images. These images and their annotations can then be used to train a DL model such as a Deep Convolution Neural Network (CNN) to produce the target model. As the number of annotations can be quite high, this becomes very labor intensive and diminishes the advantages of using CNNs. 

In this work we address this problem by proposing a method for efficient collection of expert annotations for pedestrian tracking using depth field images.

Our contribution is twofold:
\begin{itemize}
\item we design a CNN model that can detect a large number of objects in a high density scenario suitable for pedestrian tracking with overhead depth field images;
\item we develop a 'soft' supervision procedure that provides training data for the model by selecting a number of patches in the original data, designing suitable transformations to the patches and generation of realistic synthetic data for the CNN model.
\end{itemize}

The obtained model can be used for real time pedestrian detection in depth field maps, possibly on large areas exploiting Graphical Processing Units (GPU).

This paper is structured as follows: in Sect.~\ref{sect:depth} we provide some selected background on depth map based crowd recording setups. On this basis, in Sect.~\ref{sect:methods} we describe our synthetic data generation procedure as well as our neural network and its training method. In Sect.~\ref{sect:perf}, we examine the detection performance. A final discussion section closes the paper.

\section{Depth-field measurements}\label{sect:depth}

\paragraph{Overhead depth-field maps for pedestrian dynamics analyses}
A typical depth-field measurement apparatus for pedestrian dynamics  include sensors placed overhead and aligned with the vertical axis. Bird eye view, in fact, avoid mutual occlusions thus eases localization tasks. Moreover, to bypass the limited range of commercial devices such as Microsoft Kinect\textsuperscript{TM}, sensors are arranged in grids to enlarge the measurement areas. This requires a merge of the individual sensor signal. Different stitching approaches have been considered: for instance, in~\cite{corbetta2016continuous}, depth images are unified into large depth frames to then undergo detection and tracking algorithms. In~\cite{seer2014kinects}, the tracking information from different sensors are merged \textit{a posteriori}. In the next paragraph, we review some elements of the former approach as it supports the depth maps ``combination algebra''  we employ to generate synthetic annotated  depth maps (see Sect.~\ref{sect:synthannot})

\paragraph{Depth-field maps combination algebra}
In~\cite{corbetta2016continuous}, depth-field maps from neighboring Kinect sensors are merged enabling to track pedestrians over a relatively large area. This requires two operations:
(1) the depth-field measured from each sensor is converted from a perspective view into an axonometric view. Overhead axonometric views of pedestrians are translation invariant, which means that a pedestrian is represented with the same depth patch regardless whether they are at the center or at the edges of the sensor view (cf. patches in Fig.~\ref{fig:pedClust}).
(2) The axonometric view enables seamless combination of depth-field maps from neighboring sensors. Given two neighboring sensors returning the depth maps $d_1 = d_1 (x,y)$ and  $d_2 = d_2 (x,y)$ we combine them as
\begin{equation}\label{eq:depth-comp}
c((d_1,\tau_1),(d_2,\tau_2)) = \mbox{\textit{minimum}}\{(d_1 \circ \tau_1),  (d_2 \circ \tau_2 )\}.
\end{equation}
In words, the depth field maps are first rigidly translated (by a composition with the motions $\tau_1$ and $\tau_2$),  to register with the relative positions of the sensors in the physical space. Then the component-wise minimum is extracted. This is used to retain for each location the value of smallest distance, actually observed from an aerial view.
 The operation in Eq.~\eqref{eq:depth-comp} is commutative and associative, so it can be extended to an arbitrary number of cameras (or depth patches). 

\section{Method}\label{sect:methods}
\subsection{CNN model for object localization}\label{sect:cnnmethod}
Neural networks, more particularly CNNs have demonstrated particular success in image analysis~\cite{russakovsky2015imagenet}. The major advantage is their hierarchical structure that allows the models to build complex features and form efficient representation of the input data. In this method we aim to leverage this advantage to improve the detection and localization of pedestrians from depth maps. We expect that efficient features and sufficient supervision will produce models that better disentangle multiple nearby objects and make more accurate distinction between pedestrian and other objects in the scene.   

The architecture of the proposed CNN model is closely related to the YOLO object localization approach~\cite{Redmon_2016_CVPR}. This approach offers computationally efficient localization. This opens the possibility for real-time analysis or large number of objects, which are advantageous properties for large scale pedestrian tracking. 

The model processes the whole image in a single pass and produces a set of bounding boxes for each object that it has detected in the image. The model can also associate a class to each object. For our application we only provide detection of the objects, since we do not need to detect different types of objects. 
The model overlays a grid over the image, and produces a binary detection decision for each cell in the grid. It also produces an offset and the width and height of the bounding box for that object. The bounding box size is not limited to the cell size. 
For a  $S\times S$ regular grid the model produces:
\begin{equation*}
\textbf{z}_i = (x_i,y_i,w_i,h_i,n_i,p_i)\quad \mbox{with}\ \  i=1,\ldots,S^2,
\end{equation*}
where $x_i,y_i$ are the Cartesian coordinates of the  bounding box at the $i$-th tile, whose width and height are, respectively, $w_i$ and $h_i$. $p_i$ denotes the probability that $\textbf{z}_i$ is actually  a bounding box. Namely, for ground truth data, $p_i = 0$ whenever $\textbf{z}_i$ plays the role of a placeholder, conversely $p_i = 1$ states that the $i$-th is a non void bounding box. Finally, $n_i = 1 - p_i$ and is kept for extensibility to multi-type object detection.
\begin{figure*}[t]
  \begin{center}
    \includegraphics[width=0.95\linewidth]{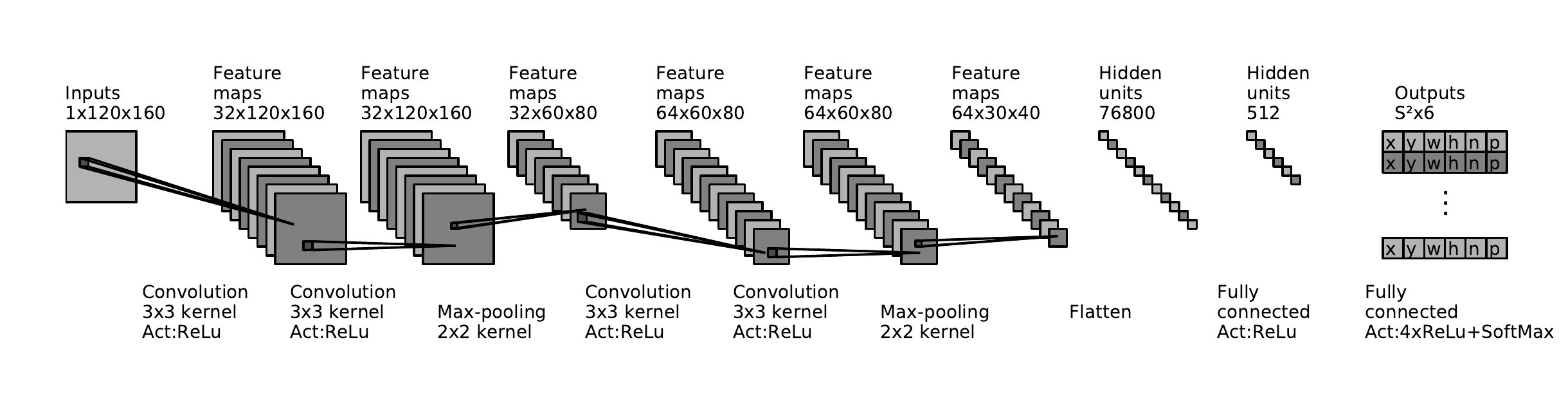}
  \end{center}
  \caption{Diagram of the neural network employed. Kernel sizes and activation functions (Act) are below each layer. }
  \label{fig:NN}
\end{figure*}

The network we employ is composed of a first section aimed at feature extraction. This is followed by a densely connected section that combines local features into bounding boxes estimates. The input are depth images (thus single channel) at resolution $160\times 120$  (VGA resolution  downsampled by a factor $4$ in each direction). Feature extraction occurs through two stacked layer blocks, each of which contains two convolutional layers with small filter size ($3\times 3$) and a final max pool layer. This architecture is closely related to the VGG network~\cite{simonyan2014very}. The convolutional layers and pooling layers are followed by fully connected layers that end with the output layer consisting of linear outputs for the bounding box parameters and softmax for the detection probabilities. The diagram of the network is given in Fig.~\ref{fig:NN}.

As non-linear function approximators, DL models are trained through a non-convex optimization procedure that minimizes a defined loss function. Due the complex multi-part output of our model, we needed to define multi-part loss function $L$: 
\begin{equation}
L
= \sum_{i=1}^{S^2}\lambda_{\mathcal{H}} \tilde{\mathcal{H}}(\textbf{z}_i,\textbf{z}_i^{gt}) + \lambda_{\mathcal{L}_2}p_i^{gt}\tilde{\mathcal{L}}_2^2(\textbf{z}_i,\textbf{z}_i^{gt}),
\end{equation}
Respectively, it holds
\begin{itemize}
\item the ``$gt$'' denotes a ground truth (synthetic) bounding box data;
\item $\tilde{\mathcal{H}}$ is the categorical cross-entropy function restricted to the $(n_i,p_i)$ components of the bounding box vector $\textbf{z}_i$;
\item $\tilde{\mathcal{L}}_2$ is the ordinary Euclidean distance among the spatial components  $(x_i,y_i,w_i,h_i)$  of the bounding box vector. Notably, the function is multiplied by $p_i^{gt}$, which acts as a switch, turning off the loss for the location parameters when there is no object present in the ground truth;
\item $\lambda_{\mathcal{H}}$ and $\lambda_{\mathcal{L}_2}$ are weighting factors for the linear combination of the two metrics.
\end{itemize}

\subsection{Weak supervision through synthetic data}\label{sect:synthannot}
DL methods rely on training data to develop models. Beyond this and the network architecture, most of the options for adding expert knowledge are indirect. One common way to guide the training is to augment or add synthetically generated training data. This allows for adding properties to the model such as invariances to translation, rotation, mirroring and skewing. One can even go further and add noise or synthetic generation of data that relies on understanding of the domain. This way the model is exposed to larger range of variances from the input space and can achieve better generalization with smaller amount of natural data available.  
We use this opportunity to deal with the difficulty in providing annotations for our problem. We achieve this by selecting patches from the original data that correspond to pedestrians. The patches provide for both a example of a pedestrian and an annotation of the bounding box around the pedestrian. We then use these patches and other patches of non-pedestrian objects to build a synthetic image for training. 

We further inject expert knowledge about the the real data by deciding on how the synthetic images are composed and by applying carefully designed transformations to the selected patches. In this manner we achieve an outstanding amount of training data will little effort from the expert.  
			
\begin{figure}[t]
  \begin{center}
    \includegraphics[width=0.7\linewidth]{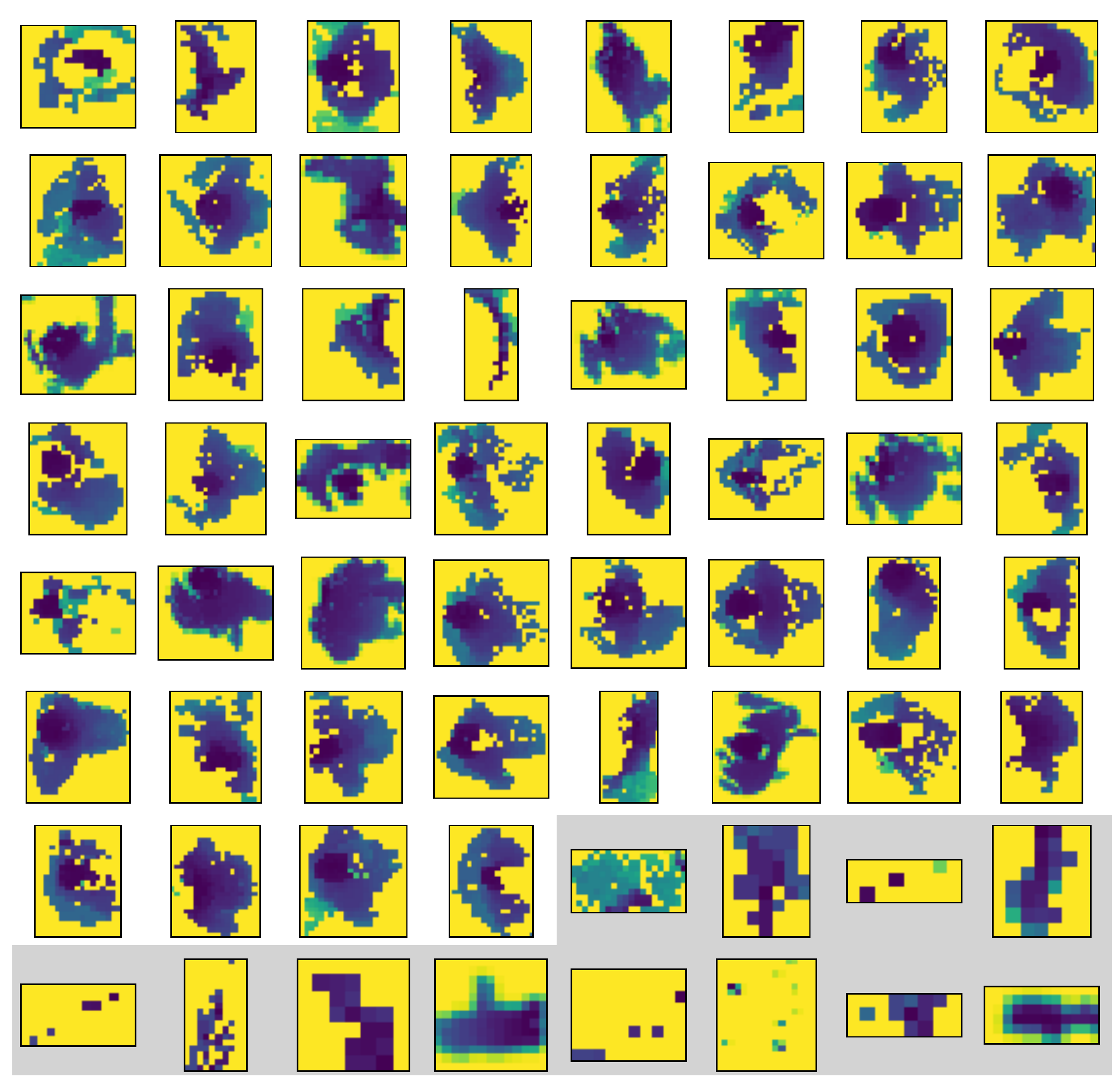}
  \end{center}
  \caption{A selection of patches defined by the expert and used for the synthetic data generation. The last $12$ items are depth artifacts or objects to be ignored by the localization algorithm.}
  \label{fig:patches}
\end{figure}

\begin{figure}[t]
  \begin{center}
    \includegraphics[width=0.95\linewidth]{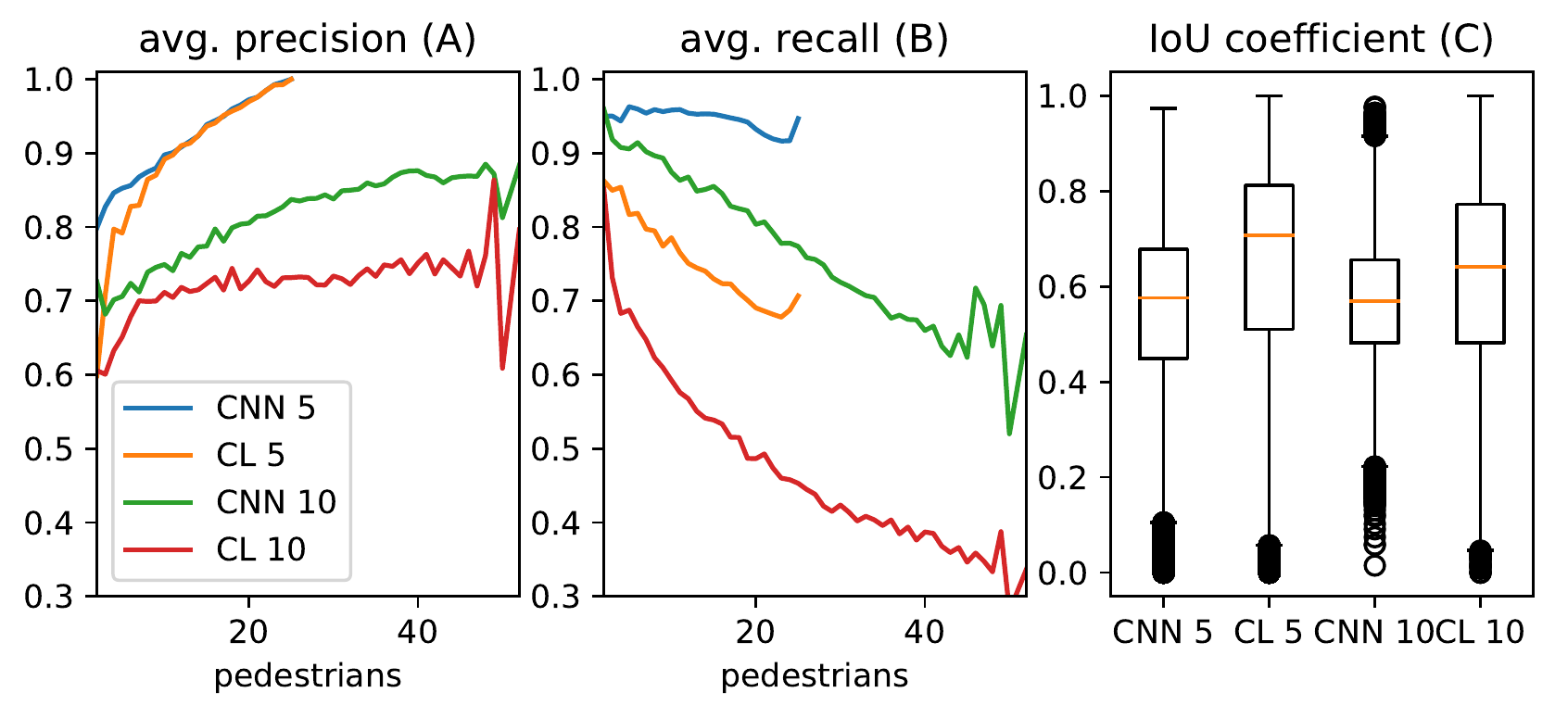}
  \end{center}
  \caption{Comparison between CNN and CL localization approaches in case $S=S_a=5$ and $S=S_a=10$. We include average precision (A) and average recall (B), both conditioned to the number of pedestrians observed, and intersection over union  coefficient (C).}
  \label{fig:Comp-quant}
\end{figure}

\begin{figure*}[t]
  \begin{center}
\includegraphics[width=0.9\linewidth]{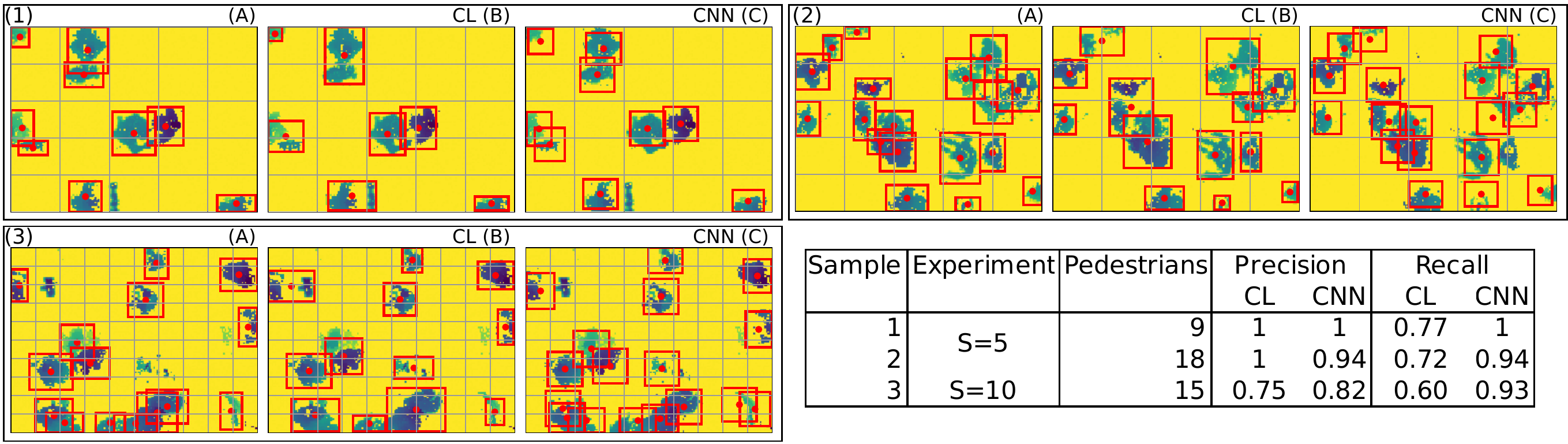}
  \end{center}
  \caption{Localization algorithms in action in our two experiments. Comparison of synthetic data (A),  clustering-based algorithm (B), and our CNN (C). Case (1) is analogous to Fig.~\ref{fig:pedClust}. Precison and recall data for the three samples are in the table. A higher recall for the CNN approach can be observed as side by side subjects and static objects are typically correctly recognized.    
  }
  \label{fig:Comp-visual}
\end{figure*}

More specifically the approach relies on a human expert to identify few hundred bounding boxes among those annotated correctly by a clustering-based algorithm (cf. Sect~\ref{sect:intro}). Here we employ a random selection of real depth maps from existing crowd tracking experiments. These are typically result of the combined output of multi-sensor setups. This a set of ``overhead human patches'' $P_u$. 

Secondly, we ask the human expert to manually extract patches that are not pedestrians. These can be of two types: 
\begin{inparaenum}
\item objects and architectonic elements (such as bags, strollers, carts, tables, and doors), $P_o$;
\item depth artifacts from sensor errors (noisy ``stain-looking'' patches in the depth field, counting from few pixels to few dozens), $P_n$.
\end{inparaenum}

We combine via Eq.~\eqref{eq:depth-comp}  elements randomly extracted from augmented versions of $P_u$,  $P_o$ and $P_n$, say $\hat P_u$,  $\hat P_o$ and $\hat P_n$. We begin with an empty depth map at VGA resolution (native output resolution of single Kinect sensors), on which we overlay a $S\times S$ regular grid (cf. network output in Sect.~\ref{sect:cnnmethod}). For each grid tile, we choose, with probability $q$ whether to place an additional  pedestrian  patch chosen randomly from $\hat P_u$. We assign to the patch centroid a position on the tile surface with uniform probability.  Thus, the total number of pedestrians in the depth maps follows a binomial distribution with $n=S^2$ and probability $q$. As a second step, we extract $N\sim\mbox{\textit{Poisson}}(\lambda)$ patches from $\hat P_o \cup \hat P_n$, that we place at random in the final map. 

We further apply random transformation to the patches including: $90^o$ rotations, flipping, pixel removal (i.e. pixels are replaced with the floor depth) and addition (i.e. pixels are replaced with the median value of the image), and rigid depth translation. Finally, we add Gaussian noise to the produced image as another layer of regularization.
 
\section{Experiment design}\label{sect:experiment}
        
We conduct two experiments aiming at comparing the performance of our CNN with a CL algorithm. The experiments differ in the grid size adopted for the CNN, respectively $S=S_a=5$ and $S=S_b=10$. Each of our CNN undergoes a training phase against synthetic data (cf. Sect~\ref{sect:synthannot}). Then, we expose both the CNN and the considered CL algorithm to further synthetic data (more general and not seen during the training) and we measure their performance.
        
\textbf{Performance evaluation} To grade the performance of the algorithms we consider their output on a cell basis. We evaluate the algorithms \textit{precision} (i.e. $tp/(tp+fp)$, cf. explanation below) and \textit{recall} (i.e. $tp/(tp+fn)$). We account cell measurement as a true positive, $tp$, if the cell is correctly predicted to hold the centroid of a bounding box (let $fp$ and $tn$ denote, respectively, false positives  and true negatives).  We consider the CNN output $\mathbf{z}_i$ to hold a bounding box prediction if $p_i > 0.5$. We further score the accuracy of each true positive computing the \textit{intersection over union} (IoU) of predicted and actual bounding box. We keep the number of pedestrians as a parameter in the analysis, as we expect it to be a major determinant of performance degradation.

\textbf{Data and CNN training}
We employ pedestrian patches, as in Fig.~\ref{fig:patches}, extracted from past measurement setups in which the sensor was located approximately $4\,$meters above the ground.  Similarly to a single Kinect operating in these conditions, synthetic depth-field maps cover an area of about\footnote{Calculated considering the characteristic field of view of Kinect sensors~\cite{Kinect} plus the assumption that individuals of height larger or equal than $1.4\,$m have to be fully resolved, even at the edges of the camera sight cone} $L_M\times L_m = 2.9\,m \times 2.2\,m = 6.4\,$m$^2$.  

During the training phase, we expose the two networks to synthetic data featuring a cell occupation probability $q$ respectively of $q_a=0.5$ and $q_b=0.2$. Hence, in case $S=S_a$, the network can detect up to $25$ pedestrians and, due to the grid constraint, the minimum distance admitted between the centroids of first neighbors is, in the worst case, $L_M/S_a = 0.58\,$m. During the training the network is exposed to an average of $n=12.5$ pedestrians per depth image (average density: $1.9\,$ped/m$^2$). In case $S =S_b$ the network can detect up to $100$ pedestrians and the minimum distance between the centroid of first neighbors is $0.28\,$m (this potentially encompasses people walking hand in hand). During the training the network is exposed to depth images including an average of $n=20$ pedestrians (average density: $3.1\,$ped/m$^2$). We implemented and trained our networks through the Keras library~\cite{chollet2015keras} with tensorflow GPU backend. We trained the two networks for a total of 300 epochs, each including $64.000$ random training depth maps and $6.400$ random validation depth maps (batch size: 64).

\textbf{Clustering algorithm and test data} The CL algorithm we compare with is similar to what employed in~\cite{corbetta2016continuous,seer2014kinects}. First, foreground blobs are randomly sampled. Hence, the samples undergo a complete-linkage hierarchical clustering. The clustering tree is cut at a \textit{cutoff} height comparable with the average human shoulder size. Finally, the cluster larger than a threshold are retained as pedestrians.

We compare, employing synthetic data, our two CNNs with one CL algorithm, which has fixed parameters. Synthesized test depth data include between 1 and 20 pedestrians (roughly uniformly distributed), for comparison with the first CNN, and between 1 and 35 for comparison with the second CNN.

\section{Results and discussion}\label{sect:perf}
In Fig.~\ref{fig:Comp-quant} we report the results of the experiments. Our CNN shows a precision equal (case $S=5$) or higher (case $S=10$) than the CL approach (Fig.~\ref{fig:Comp-quant}(A)), and significantly better results for recall (Fig.~\ref{fig:Comp-quant}(B)). For those cells in which a bounding box has been correctly localized, we evaluate the localization accuracy by measuring the IoU coefficient. In this case, the CNN performance is comparable with the CL approach. In Fig.~\ref{fig:Comp-visual} we include samples of synthetic depth field maps from our experiments for visual inspection. 

Following our expectations, the CNN delivers higher localization performance than the CL approach. In fact, the CNN succeeds in disentangling neighboring pedestrians and in avoiding non-pedestrian elements (cf. Fig.~\ref{fig:Comp-visual}). This results in the higher recall performance. Notably, we leveraged expert knowledge efficiently to extract the patches combined in the synthetic depth maps. As the localization quality is determined by the examples rather than hand-crafted procedures, we expect the method to generalize and transfer across different real-life measurement setups. Possibly with the only effort of enriching the patch library with object characteristic to specific location.

\section{Consclusions}\label{sect:conclusions}
We targeted a high performance pedestrian localization tool for overhead depth field data capable or running in real-time setting. Depth data is often employed in real-life pedestrian dynamics research, and, so far, hierarchical clustering-based approaches have been mostly used for localization. Here we aimed at bypassing the typical shortcomings of such approaches, leveraging the generalization power of deep learning models.

We presented a convolutional neural network approach showing significantly higher recall performance than clustering methods. This was possible as the network learns the typical shape of individuals. As such it can disentangle neighboring subjects and diminish false positive outputs (e.g. objects, depth artifacts). To bypass the difficulties imposed by the need for a large number of annotated examples, we developed a procedure producing synthetic training data as a means for efficient delivery of 'soft' supervision from an expert.  

{\small
\bibliographystyle{ieee}
\bibliography{egbib}
}


\end{document}